\documentclass[conference]{IEEEtran}
\IEEEoverridecommandlockouts
\usepackage{cite}
\usepackage{amsmath,amssymb,amsfonts}
\usepackage{algorithmic}
\usepackage{graphicx}
\usepackage[table,xcdraw]{xcolor}
\usepackage{textcomp}
\usepackage{xcolor}
\def\BibTeX{{\rm B\kern-.05em{\sc i\kern-.025em b}\kern-.08em
    T\kern-.1667em\lower.7ex\hbox{E}\kern-.125emX}}
\begin{document}

\title{Seamless Integration: Sampling Strategies in Federated Learning Systems\\
}
\makeatletter
\newcommand{\linebreakand}{%
  \end{@IEEEauthorhalign}
  \hfill\mbox{}\par
  \mbox{}\hfill\begin{@IEEEauthorhalign}
}
\makeatother
\author{\IEEEauthorblockN{1\textsuperscript{st} Tatjana Legler}
\IEEEauthorblockA{ 
\textit{University of Kaiserslautern-Landau (RPTU)}\\
\textit{German Research Center for Artificial Intelligence (DFKI)}\\
Kaiserslautern, Germany \\
0000-0002-7297-0845}
\and
\IEEEauthorblockN{2\textsuperscript{nd} Vinit Hegiste}
\IEEEauthorblockA{
\textit{University of Kaiserslautern-Landau (RPTU)}\\
Kaiserslautern, Germany \\
0000-0001-6944-1988}
\linebreakand 
\IEEEauthorblockN{3\textsuperscript{rd} Martin Ruskowski}
\textit{University of Kaiserslautern-Landau (RPTU)}\\
\textit{German Research Center for Artificial Intelligence (DFKI)}\\
Kaiserslautern, Germany\\
0000-0002-6534-9057}

\onecolumn
© 2024 IEEE. Personal use of this material is permitted. Permission from IEEE must be obtained for all other uses, in any current or future media, including reprinting/republishing this material for advertising or promotional purposes, creating new collective works, for resale or redistribution to servers or lists, or reuse of any copyrighted component of this work in other works.

Accepted to be published in: The 2nd IEEE International Conference on Federated Learning Technologies and Applications (FLTA24)
\twocolumn

\maketitle

\begin{abstract}
Federated Learning (FL) represents a paradigm shift in the field of machine learning, offering an approach for a decentralized training of models across a multitude of devices while maintaining the privacy of local data. However, the dynamic nature of FL systems, characterized by the ongoing incorporation of new clients with potentially diverse data distributions and computational capabilities, poses a significant challenge to the stability and efficiency of these distributed learning networks. The seamless integration of new clients is imperative to sustain and enhance the performance and robustness of FL systems. This paper looks into the complexities of integrating new clients into existing FL systems and explores how data heterogeneity and varying data distribution (not independent and identically distributed) among them can affect model training, system efficiency, scalability and stability. Despite these challenges, the integration of new clients into FL systems presents opportunities to enhance data diversity, improve learning performance, and leverage distributed computational power. In contrast to other fields of application such as the distributed optimization of word predictions on Gboard (where federated learning once originated), there are usually only a few clients in the production environment, which is why information from each new client becomes all the more valuable. This paper outlines strategies for effective client selection strategies and solutions for ensuring system scalability and stability. Using the example of images from optical quality inspection, it offers insights into practical approaches. In conclusion, this paper proposes that addressing the challenges presented by new client integration is crucial to the advancement and efficiency of distributed learning networks, thus paving the way for the adoption of Federated Learning in production environments.
\end{abstract}


\section{Introduction}

Federated Learning (FL) has emerged as a transformative approach in the field of machine learning, fundamentally changing how data is utilized and models are trained across distributed networks \cite{BrendanMcMahan.2017}. In contrast to traditional centralized machine learning paradigms, FL allows numerous devices to collaboratively train a shared model while retaining all training data locally on the devices. This approach addresses significant concerns related to data privacy, security, and access rights \cite{Kairouz.2021}. The decentralized nature of FL not only reduces the risks associated with data centralization but also enables the use of data that was previously inaccessible or unusable due to privacy constraints.

The applications of machine learning in manufacturing span a wide range, from well-established uses such as fault diagnostics, process monitoring, and anomaly detection to more recent innovations like enhancing human-robot collaboration with the aid of Generative Pre-trained Transformers (GPTs) and optimizing construction designs \cite{Nti.2022}. The demand for efficient data processing, coupled with the sensitive and proprietary nature of manufacturing data, presents numerous challenges for machine learning applications \cite{Dogan.2021}. These challenges are exacerbated in the dynamic environment of FL, particularly with the continuous integration of new clients. These new participants, characterized by potentially diverse data distributions and varying computational capabilities, introduce complexities that can impact the stability, efficiency, and scalability of FL systems. The heterogeneity in data, often non-independent and identically distributed (non-IID), along with the variance in computational resources among clients, necessitates innovative strategies to seamlessly incorporate these new participants into existing networks \cite{Arafeh.2022} \cite{Li.2020b}.

The introduction of new clients within a FL system offers the potential to enhance the diversity of the data pool, thereby improving the robustness and performance of the collective learning process. Each new client contributes unique insights and perspectives, enriching the network and making the model more comprehensive and better suited for real-world applications. This is especially valuable in fields like optical quality inspection, where variations in data across different production lines can significantly influence the effectiveness of predictive models.

To fully harness the potential benefits of FL, it is crucial to address the challenges related to the integration of new clients. This requires the use or development of strategies that enable effective onboarding and maintain the stability of the FL system. In this paper, we focus on client selection techniques as a foundation for future research into client integration, examining the complexities introduced by data heterogeneity and proposing strategies for efficient client integration.

For decades, various approaches, such as zero-defect manufacturing and lean Six Sigma methodologies, have been employed in production environments \cite{Myklebust.2013}\cite{Hassan.2013}. These methods have successfully reduced defect rates to levels measurable in "parts per million" \cite{Calvin.1983}. However, with the increasing variety of product variants, these traditional methods are reaching their limits \cite{Berry.1999}. To maintain quality assurance amidst this diversity, machine learning methods are increasingly being applied \cite{Rai.2021}.

Several strategies exist to overcome data scarcity and expand the usable data pool. Generative AI methods, for example, can create synthetic data for multiple product variants and configurations, including training data for various error cases \cite{Buggineni.2024}. Additionally, breaking down existing data silos can increase both the volume and diversity of available data \cite{Adams.2023}. In the following, we explore FL as a potential approach to achieve this while preserving data privacy.  Additionally, synthetic data and real images can be merged with FL, enhancing the robustness and applicability of machine learning models in production, as shown in \cite{Hegiste.2024}.

\section{State of the Art}
The manufacturing industry is a data-intensive sector with numerous actors generating data from various sources, including sensors, production processes, and quality control systems \cite{Liu.2020}. However, the data in manufacturing are often sensitive and protected, complicating the sharing or collaborative training of machine learning models on a shared dataset \cite{Dogan.2021}. FL offers a decentralized approach that aims to leverage a broad, heterogeneous database without the need to share raw data \cite{Kairouz.2021}. Unlike traditional centralized learning, where models are trained on large datasets housed on a central server \cite{Dean.2012}\cite{Iandola.2016}, FL only shares model parameters, such as network weights, rather than raw information like production data or images. This approach allows each participant to benefit from data-sovereign learning while reaping the advantages of collaborative learning, such as improved generalizability and stability. Additionally, by indirectly increasing the sample size, FL can provide access to a greater variety of features.

Beyond the basic concept of FL, McMahan et al. introduced a detailed method called Federated Averaging (FedAvg), in which a central server functions as an aggregator \cite{McMahan.2017}. The FedAvg process includes the following steps:

\begin{itemize}
\item \textbf{Initialization:} A server initializes a global model and distributes it to all participating clients, initiating their local training.

\item \textbf{Local Training:} Each selected client trains the model on its local data.

\item \textbf{Aggregation:} After local training is completed on the device, the model updates (e.g., weights or weight changes) are sent back to the central server. The server then aggregates these updates to create the global model, typically by calculating an arithmetic average, as done in FedAvg.

\item \textbf{Update:} The global server sends the updated model to all clients.
\end{itemize}

This process, known as a communication round, is repeated until the model achieves the desired accuracy or meets a specific convergence or termination criterion (see Figure \ref{fig:FL}).

\begin{figure*}
    \centering
    \includegraphics[width=0.75\textwidth]{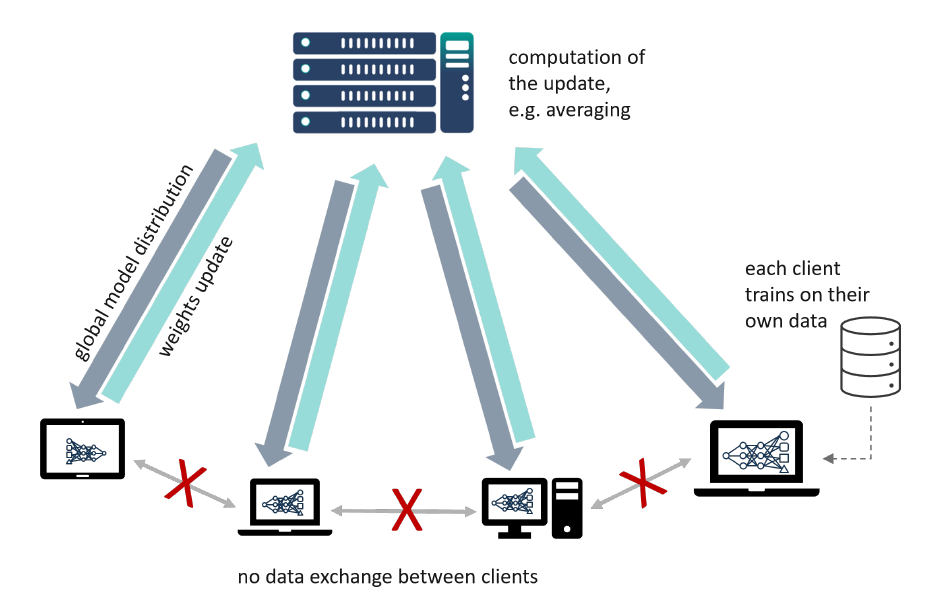}
    \caption{Illustration of the federated learning process \cite{Legler.2023}.}
    \label{fig:FL}
\end{figure*}

\subsection{Client Selection Strategies}
When FL and FedAvg were introduced, initially a \textit{random selection} of clients was performed to participate in the next communication round \cite{McMahan.2017}. The authors observed that incorporating additional clients beyond a certain threshold yields diminishing returns as the additional communication overhead exceeds the performance gains, but did not investigate further into client selection strategies. This simple approach works well when there is a sufficient number of clients with similar data distributions \cite{Li.2019}. However, it often fails to account for device heterogeneity, which can lead to incomplete communication rounds, or the risk of over-sampling frequently provided data \cite{Cho.2020}.

The available computation power, the network connection as well as a possible battery operation which can lead to phases of energy saving during which the clients are inaccessible, can be considered for \textit{resource-aware selection} \cite{Maciel.2023}. Using different hardware components can result in different calculation times. For instance,  \cite{Tahir.2022} focuses on reducing the influence of hardware in the selection process by neither waiting for the slowest clients (known as stragglers) every round nor completely discarding them, which would result in an imbalance in representation. To avoid a prolonged waiting phase in mobile edge computing as devices with varying resources communicate over a cellular network, \cite{Nishio.2019}  estimates how many clients can reasonably participate.

For \textit{performance-aware selection} metrics such as accuracy or training loss can be used in addition to the consideration of resource constraints as mentioned above \cite{Panigrahi.2023}. \cite{Cho.2020b} demonstrates, that the choice of clients with a higher local loss can improve convergence of the model, as they have a bigger potential for improvement.

The term \textit{Contribution-Based Selection}  is used to categorize methods that use the effects of a client on the global model as a selection criterion for the next communication round \cite{Lin.2022}. \cite{Qiao.2023} uses the Shapley value known from cooperative game theory to estimate the contribution of each client to the global model. A further development can be found in \cite{Jiang.2023} resulting in a contribution estimation, which looks at the gradient space and data space for each client. \cite{Ruan.2021} propose a flexible selection scheme that converges even when communication rounds are interrupted by clients leaving or joining. 

\cite{Hegiste.2024b} investigate weight selection strategies in federated learning, focusing on scenarios with few clients and repetitive product images in local datasets to mirror real-world industrial conditions. Their comprehensive evaluation across multiple deep learning architectures reveals that Optimal Epoch Weight Selection (OEWS) consistently outperforms other strategies, significantly boosting model performance in industrial applications.

\textit{Clustering} approaches can be applied in various contexts within FL. Clustering can be differentiated based on the input parameters used, such as training weights, known information like geolocation data, computed metrics such as accuracy, data distribution details, or resource availability. The objectives of clustering can also vary. Some methods aim to form clusters for further training as separate groups, thereby improving local accuracy. Others use clustering to select a diverse set of clients for the next training round, with the goal of developing a more robust global model.

\cite{Arisdakessian.2023} proposes a multi-criterial clustering approach, where the task owner can choose resource-based (such as CPU or RAM) or data-oriented (such as data quality or minimal size) requirements. They show that their DBSCAN-based selection of clients performs better than randomly selecting the same number of clients. One drawback with their approach on the one hand is that they rely on the correctness of sent information from clients regarding resources and data and that it has only been tested at MNIST.  

\cite{Li.2022} cluster the clients based on the label of the local faults of a computer room and use the Hamming distance to determine similarity. As the Hamming distance describes the number of symbols that distinguish two strings of equal length from each other \cite{Hamming.1986}; this method is only useful if all possible error cases are predefined and displayed in a binary representation. This method is not suitable for recognizing similarities from only the underlying data or weights.


An initial Clustered Federated Learning (CFL) approach was introduced in \cite{Sattler.2021}, where clients are iteratively bipartitioned based on the cosine similarity of gradient changes. Hierarchical clustering offers the advantage of not requiring the number of clusters to be known in advance \cite{Johnson.1967}, and it also provides scalability and interpretability of the results. However, since multiple communication rounds are needed to fully separate all clients, recursive bipartitioning clustering demands significant computational and communication resources, which can limit its practical feasibility in scenarios with many participants. Additionally, because clients are randomly selected each round, the algorithm can be computationally inefficient and may even completely fail, further restricting its practical applicability \cite{Duan.2021}.

\cite{Briggs.2020} also employs hierarchical clustering to divide clients based on the similarity of gradient changes, but performs the clustering after a predetermined number of communication rounds. The procedure begins by training a global model, which is then fine-tuned on the private datasets of all clients to assess the difference between the global and local model parameters. This difference is used to create clusters and assign clients to them. Subsequent training is conducted independently within each cluster, resulting in the creation of multiple models. This approach is designed to accommodate a wider range of non-IID settings and allows for training on a subset of clients during each round of the FL model training.

To improve the efficiency of CFL, \cite{Ghosh.2020} proposes the Iterative Federated Clustering Algorithm (IFCA), which randomly generates cluster centroids and assigns clients to clusters in a way that minimizes the loss function. Although model accuracy with IFCA is significantly improved compared to conventional FL, the communication costs remain unchanged, and the computational effort is higher, as all updates must be transmitted and clusters reformed in each round. Additionally, the probability of success is highly dependent on the initialization of the cluster centroids.

Cosine similarity followed by affinity propagation clustering has been shown to effectively cluster clients in image classification settings, as demonstrated by \cite{Tian.2022}. Their evaluation, conducted on the MNIST and CIFAR-10 datasets, included a verification procedure to ensure that the new global weights do not reduce accuracy below a predetermined threshold. If the accuracy drops below this threshold, the new weights are rejected, and training continues with the local weights.

Many purely academic approaches often yield effective solutions on benchmark datasets and demonstrate the convergence of their methods to address specific problems \cite{Beutel.2020}. These problems are typically framed as closed systems, where the primary objective is to optimize specific metrics, such as achieving high accuracy or minimizing the number of communication rounds. However, this perspective frequently overlooks the dynamic nature of real-world systems, which can operate over extended periods. In practice, clients may join or leave the system at any time, or introduce entirely new data, adding layers of complexity that are not fully accounted for in controlled, static environments.

For our further considerations, we will disregard the hardware limitations of the clients, as it is assumed that companies have sufficient computing resources and stable network connections. In this context, optimizing model performance takes precedence over the duration of communication rounds. However, we will ensure that the required computational power remains reasonable for an industrial environment. Additionally, we assume a cross-silo problem definition, characterized by a relatively small number of clients \cite{Kairouz.2021,Huang.2021}. In this paper, we will focus on examining various client selection techniques to lay the groundwork for improving the integration of new clients in future work. This is particularly important because the careless integration of new clients into a stable system can lead to unwanted performance drops until the system stabilizes again with new clients. Our future work will evaluate how effectively these techniques can incorporate new clients, along with their data and data distributions, into the system.

\section{Methodology}
First, we examined how clustering approaches, which were primarily demonstrated on MNIST and minimal examples, can be transferred to real-world datasets and state-of-the-art neural networks, such as EfficientNet or DenseNet. Subsequently, we evaluated various clustering methods and selected the most effective one for further analysis. Initial testing was conducted on a somewhat simpler, proprietary dataset (see section \ref{sec:Dataset}) to establish a baseline understanding, which was then validated on a subset of the more complex ImageNet dataset. This approach allowed us to assess the generalizability and robustness of the selected clustering method across different types of data and neural network architectures.

Even relatively small neural networks typically contain millions of trainable parameters. When fine-tuning on custom datasets, it is common practice to adapt only the final layers to the specific data and the required number of output classes. Consequently, analyzing only the final layer's activations is often sufficient for effective clustering, as it captures the most relevant features learned during fine-tuning. In this case, with a shape of (n, 1280) corresponding to n classes in EfficientNet, this approach results in a small sample size with high dimensionality, which presents unique challenges for clustering.

\subsection{Dataset}\label{sec:Dataset}
Our Dataset consists of cabins of miniature trucks, where the parts can be 3D printed, milled, or bought separately. See Figure \ref{fig:truck} for two exemplary images from the quality inspection module. As the product is assembled on demand in lot size one, many combinations are possible. Previous publications proved that one advantage of FL is the higher generalizability \cite{Hegiste.2022}, therefore allowing us to take many pictures away from the production line, where a higher variability of image date is easier to obtain and the risk of interfering with the production process is avoided. For testing purposes, we split the data set so that each client has only one color and one type of windshield, in addition to the class without windshield for every client. The distribution can be obtained from Table \ref{tab:distribution}: Classes 1 to 4 are evenly distributed at approximately 12\%, whereas class 0 is the predominant category, comprising nearly 50\% of the total distribution. In contrast, class 5 is underrepresented, accounting for only about 5\%.
For some tests, not all clients take part in the federation from the beginning and simulate newcomers that join with new data.
 
\begin{figure}
    \centering
    \includegraphics[width=1\linewidth]{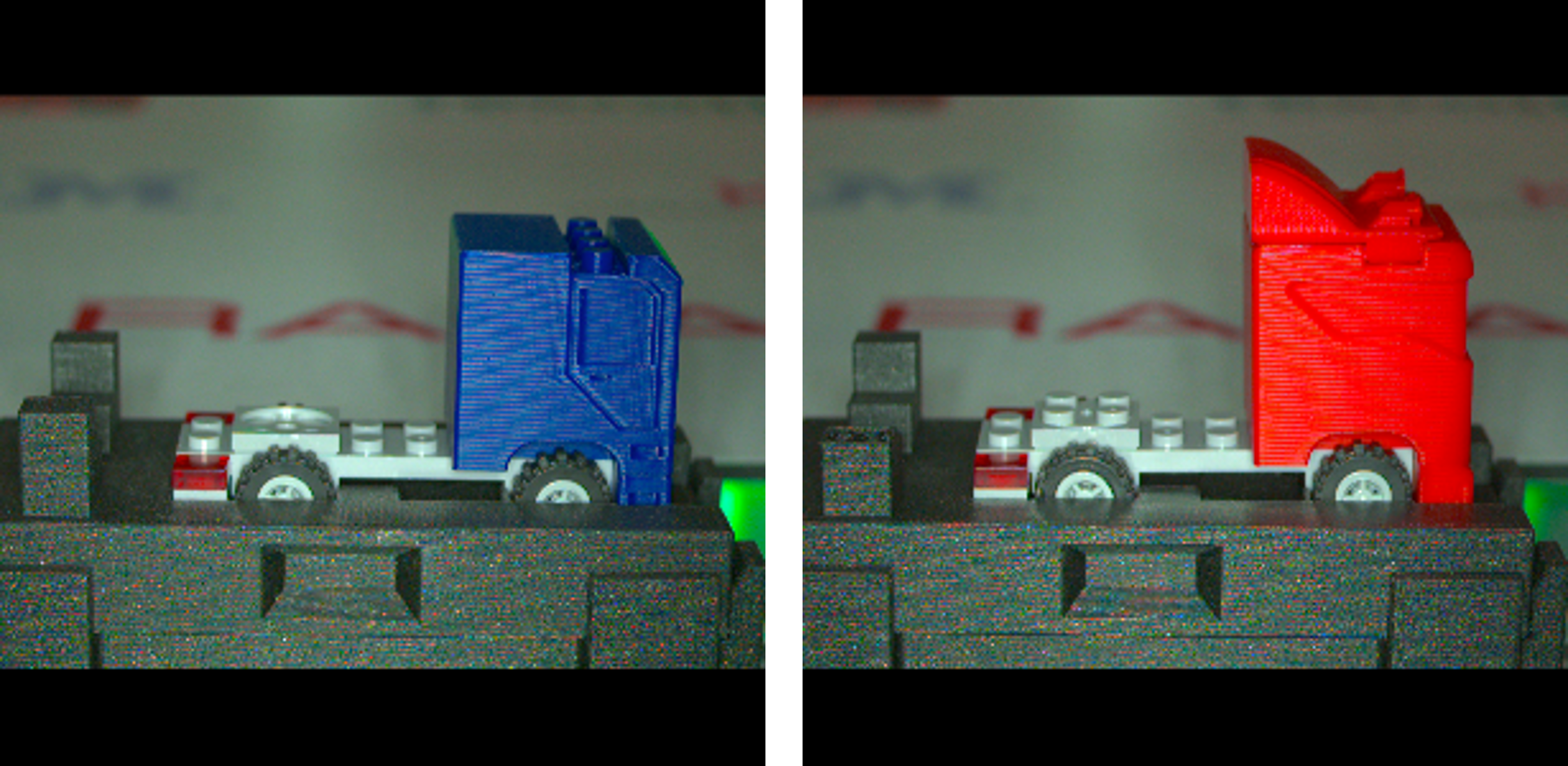}
    \caption{Miniature Truck, captured by an inline quality inspection module. Only a cabin with one type of windshield (red) is pictured. In the following step, one trailer is added from a selection of options.}
    \label{fig:truck}
\end{figure}

\begin{table}[h]
\caption{Distribution of the training data to the classes}
\label{tab:distribution}
\begin{tabular}{l|llllll}
\textbf{Class}   & \textbf{0} & \textbf{1} & \textbf{2} & \textbf{3} & \textbf{4} & \textbf{5} \\ \hline
\textbf{CID}     &            &            &            &            &            &            \\
\textbf{1}       & 565        & 659        &            &            &            &            \\
\textbf{2}       & 565        &            & 574        &            &            &            \\
\textbf{3}       & 565        &            &            & 625        &            &            \\
\textbf{4}       & 565        &            &            &            & 627        &            \\
\textbf{5}       & 499        & 499        &            &            &            &            \\
\textbf{6}       & 499        &            & 504        &            &            &            \\
\textbf{7}       & 499        &            &            & 511        &            &            \\
\textbf{8}       & 499        &            &            &            & 499        &            \\
\textbf{9}       & 528        & 565        &            &            &            &            \\
\textbf{10}      & 528        &            & 532        &            &            &            \\
\textbf{11}      & 528        &            &            & 559        &            &            \\
\textbf{12}      & 528        &            &            &            & 506        &            \\
\textbf{13}      & 510        & 525        &            &            &            &            \\
\textbf{14}      & 510        &            & 530        &            &            &            \\
\textbf{15}      & 510        &            &            & 488        &            &            \\
\textbf{16}      & 510        &            &            &            & 500        &            \\
\textbf{17}      & 565        & 607        &            &            &            &            \\
\textbf{18}      & 565        &            & 633        &            &            &            \\
\textbf{19}      & 565        &            &            & 566        &            &            \\
\textbf{20}      & 565        &            &            &            & 599        &            \\
\textbf{21}      & 547        &            &            &            &            & 580        \\
\textbf{22}      & 544        &            &            &            &            & 480        \\\hline
\textbf{Sum}     & 11759      & 2855       & 2773       & 2749       & 2731       & 1060       \\
\textbf{Percent} & 49,15\%    & 11,93\%    & 11,59\%    & 11,49\%    & 11,41\%    & 4,43\%    
\end{tabular}
\end{table}      

\begin{figure}
   \centering
    \includegraphics[width=1.0\linewidth]{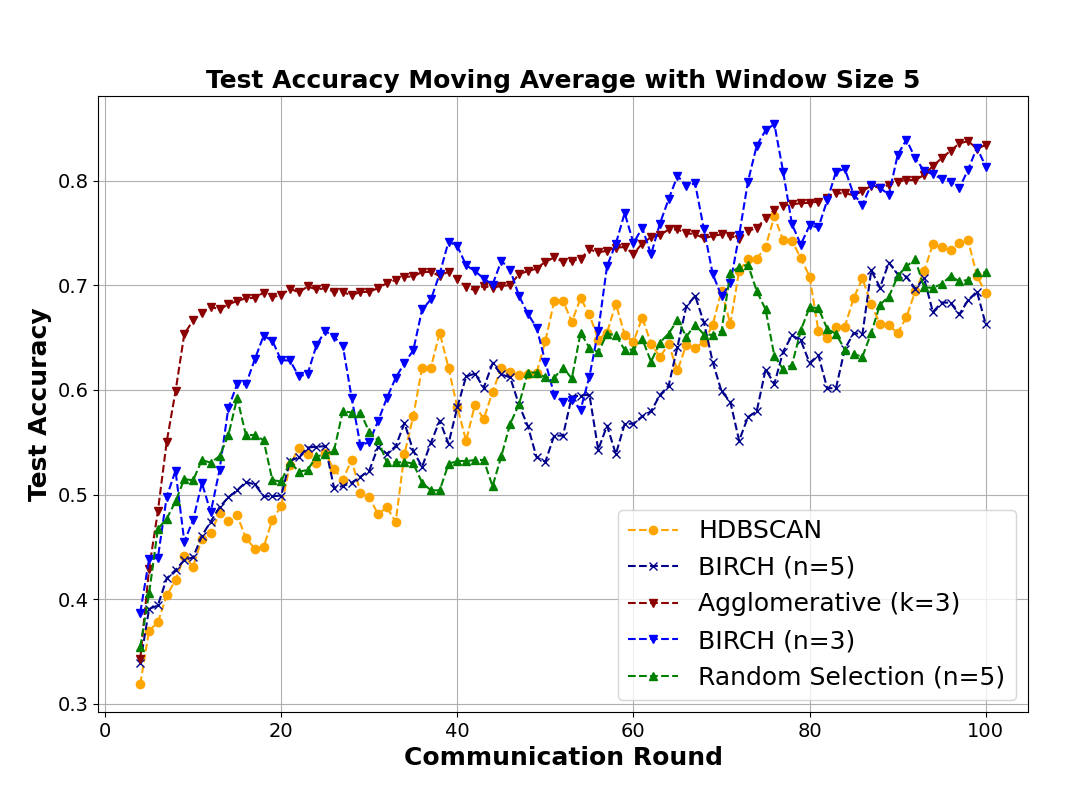}
    \caption{Comparison of clustering methods on test accuracy across 100 communication rounds.}
    \label{fig:cluster}
\end{figure}

\subsection{Usability Study on Clustering}
A pretrained EfficientNet B4 \cite{Tan.2019} was used as the frozen backbone, with one fully connected layer replaced to match the number of classes. The training was done with SGD as the optimizer with a learning rate of 0.001, 200 communication rounds and 5 local epochs. 
Using Pytorch, the preliminary examinations for clustering were carried out with our implementation to exclude any other influences (e.g. calculating the global model manually based on client's weight files).

To determine which clustering method is most appropriate, three methods that are well-suited for a large number of data points and clusters were selected: Agglomerative clustering \cite{ChidanandaGowda.1978}, HDBSCAN (Hierarchical Density-Based Spatial Clustering of Applications with Noise) \cite{McInnes.2017} and BIRCH (Balanced Iterative Reducing and Clustering using Hierarchies) \cite{Zhang.1997b}. Figure \ref{fig:cluster} presents sample results from a larger set of tests that were conducted. The optimal parameters, such as the distance metric, for each method were determined through grid search on the weights from previous runs in which all clients participated. We observe that with agglomerative clustering, accuracy increases rapidly, particularly during the first 10 rounds, and then remains stable while continuing to improve gradually. All other methods exhibit significantly higher fluctuations and, except for BIRCH (n=5), even achieve lower accuracy than agglomerative clustering. It should be noted that BIRCH occasionally even clearly surpasses it. The high fluctuation in BIRCH's performance could be due to its sensitivity to parameters and the high dimensionality combined with a small sample size. BIRCH is generally designed for larger datasets, and when applied to a small dataset with high dimensional features, it may struggle to form stable clusters, leading to the observed variability in test accuracy. HDBSCAN consistently performed worse in comparison.

Agglomerative clustering performs best in our scenario due to its ability to capture complex hierarchical structures in relatively small, but high-dimensional dataset. Its flexibility in distance metrics, robustness to small sample sizes, and stability in high-dimensional spaces make it well-suited to effectively clustering the final layer activations of your neural network, leading to more accurate and stable results compared to other methods like BIRCH or HDBSCAN.

A more detailed evaluation was caried out using saved weights from two different scenarios: one in which all clients participated in every communication round, and another where the experiment began with 8 clients (representing all classes but limited to red and blue colors), gradually adding new clients after an initial training phase. Clusters were recalculated after each communication round.
The following observations were made during the experiments: New clients were immediately recognized as newcomers, allowing for precautions to be taken to prevent any adverse effects or using advanced approaches to integrate them effectively. Within a maximum of three communication rounds, each consisting of five local epochs, new clients were correctly assigned to the appropriate cluster. It was also evident that cabin types 1 and 4 were easily distinguishable, while types 2 and 3 were often grouped together into the same cluster or formed mixed clusters.
Agglomerative clustering with complete linkage yielded good results, as did the BIRCH algorithm. However, the resulting clusters sometimes differed; for instance, BIRCH occasionally produced clusters more strongly correlated with cabin colors. In subsequent communication rounds, the clusters realigned with windshield types.
Additionally, one group exhibited a higher loss than the others, with no further improvement. Upon closer inspection of the client IDs, it became clear that the group with the higher loss consisted entirely of clients that trained with type 4 windshields. An analysis of the weights confirmed this clustering. Depending on the use case and objectives, it may be beneficial either to develop multiple models tailored to these clusters or to design the client selection process in a way that disrupts this separation, such as by consistently selecting a client from each cluster.

\subsection{Evaluation of Client Selection Strategies}
In order to compare different strategies, the clients as listed in Table \ref{tab:sampling} were selected for training and the data of the excluded clients were used for testing, whereby random images were selected to create a similar distribution of classes, e.g. to avoid over-representing class 0.
The total number of clients was 22, 16 clients correspond to the setup shown in Table \ref{tab:sampling}. In order to simulate the complete absence of a class, the last client with class 5 was omitted from the training in some tests, resulting in 15 clients. The configuration for each run was: EfficientNet with SGD as an optimizer with learning rate 0.001, 200 communication rounds with 5 local epochs each.
As a baseline, random sampling of approximately twenty and fifty percent (rounding up to the next integer number) of all participating clients was done. Between a sampling rate of 50\% and 20\% of all 22 clients, there is no noticeable difference, supporting the findings of \cite{McMahan.2017} that saturation occurs at a certain point, and additional client participation no longer results in improvement.

\begin{table}[h]
\centering
\caption{Overview of client selection for sampling tests, indicating the classes present in each client's dataset. Each 'x' represents a class the client has, while empty cells indicate missing classes. Missing classes within each color-coded group are highlighted.}
\label{tab:sampling}
\begin{tabular}{rlcccccc}
\textbf{Client ID} & \textbf{Color} & \textbf{0} & \textbf{1} & \textbf{2} & \textbf{3} & \textbf{4} & \textbf{5} \\ \hline
1 & \cellcolor[HTML]{83CCEB}blue & x & x &  &  &  & \cellcolor[HTML]{FFFFC7} \\
2 & \cellcolor[HTML]{83CCEB}blue & x &  & \cellcolor[HTML]{FFFC9E} & x &  & \cellcolor[HTML]{FFFFC7} \\
3 & \cellcolor[HTML]{83CCEB}blue & x &  &  &  & x & \cellcolor[HTML]{FFFFC7} \\
4 & \cellcolor[HTML]{FF7C80}red & x & x &  &  &  & \cellcolor[HTML]{FFFFC7} \\
5 & \cellcolor[HTML]{FF7C80}red & x &  & x &  &  & \cellcolor[HTML]{FFFFC7} \\
6 & \cellcolor[HTML]{FF7C80}red & x &  &  & \cellcolor[HTML]{FFFC9E} & x & \cellcolor[HTML]{FFFFC7} \\
7 & \cellcolor[HTML]{B5E6A2}green & x & x &  &  &  & \cellcolor[HTML]{FFFFC7} \\
8 & \cellcolor[HTML]{B5E6A2}green & x &  & x &  &  & \cellcolor[HTML]{FFFFC7} \\
9 & \cellcolor[HTML]{B5E6A2}green & x &  &  & x & \cellcolor[HTML]{FFFFC7} & \cellcolor[HTML]{FFFFC7} \\
10 & \cellcolor[HTML]{FCFF2F}yellow & x & \cellcolor[HTML]{FFFC9E}{\color[HTML]{FFCC99} } & x &  &  & \cellcolor[HTML]{FFFFC7} \\
11 & \cellcolor[HTML]{FCFF2F}yellow & x &  &  & x &  & \cellcolor[HTML]{FFFFC7} \\
12 & \cellcolor[HTML]{FCFF2F}yellow & x &  &  &  & x & \cellcolor[HTML]{FFFFC7} \\
13 & \cellcolor[HTML]{F8A102}orange & x & x &  &  &  & \cellcolor[HTML]{FFFFC7} \\
14 & \cellcolor[HTML]{F8A102}orange & x &  & \cellcolor[HTML]{FFFC9E} & x &  & \cellcolor[HTML]{FFFFC7} \\
15 & \cellcolor[HTML]{F8A102}orange & x &  &  &  & x & \cellcolor[HTML]{FFFFC7} \\
16 & \cellcolor[HTML]{ADADAD}silver & x &  &  &  &  & x \\
 &  &  & \multicolumn{1}{l}{} & \multicolumn{1}{l}{} & \multicolumn{1}{l}{} & \multicolumn{1}{l}{} & \multicolumn{1}{l}{} \\ \hline
\textbf{Sum} &  & \textbf{16} & \multicolumn{1}{l}{\textbf{4}} & \multicolumn{1}{l}{\textbf{3}} & \multicolumn{1}{l}{\textbf{4}} & \multicolumn{1}{l}{\textbf{4}} & \multicolumn{1}{l}{\textbf{1}}
\end{tabular}
\end{table}


In comparison to random selection, we evaluate metric-based methods (specifically, highest training loss) and cluster-based methods (focussing on agglomerative clustering). For the initialization of both methods, each client is selected once at the beginning.
Figure \ref{fig:hist_selection} shows the impact the methods have on client participation. A review of the selection based on the training loss reveals that client 16, with the underrepresented class, is selected with greater frequency. However, client 1, is selected with a similarly high frequency, although class 1 is represented at an average level.

Figure \ref{fig:test_acc_moving} illustrates the test accuracy over 200 communication rounds for four client selection methods: Random selection with 5 clients participating each round, loss-based selection of the 5 clients with the highest training loss, and agglomerative clustering with k=3 and k=5, corresponding to the number of clusters to be identified. The moving average with windows size 5 is used for clearer visualization. Agglomerative clustering with k=5 consistently outperforms the other methods, showing the highest and most stable accuracy throughout. Initially, it demonstrates a rapid increase in accuracy, maintaining superior performance across all phases. Agglomerative clustering with k=3 and random selection yield similar results, with k=3 exhibiting more fluctuations but eventually aligning with random selection. The training loss method, despite early instability, eventually achieves comparable accuracy, though with more fluctuations. Overall, clustering with k=5 offers the greatest accuracy and stability, while non-randomized methods provide better and more controlled integration of new clients. 

\begin{figure}
    \centering
    \includegraphics[width=1\linewidth]{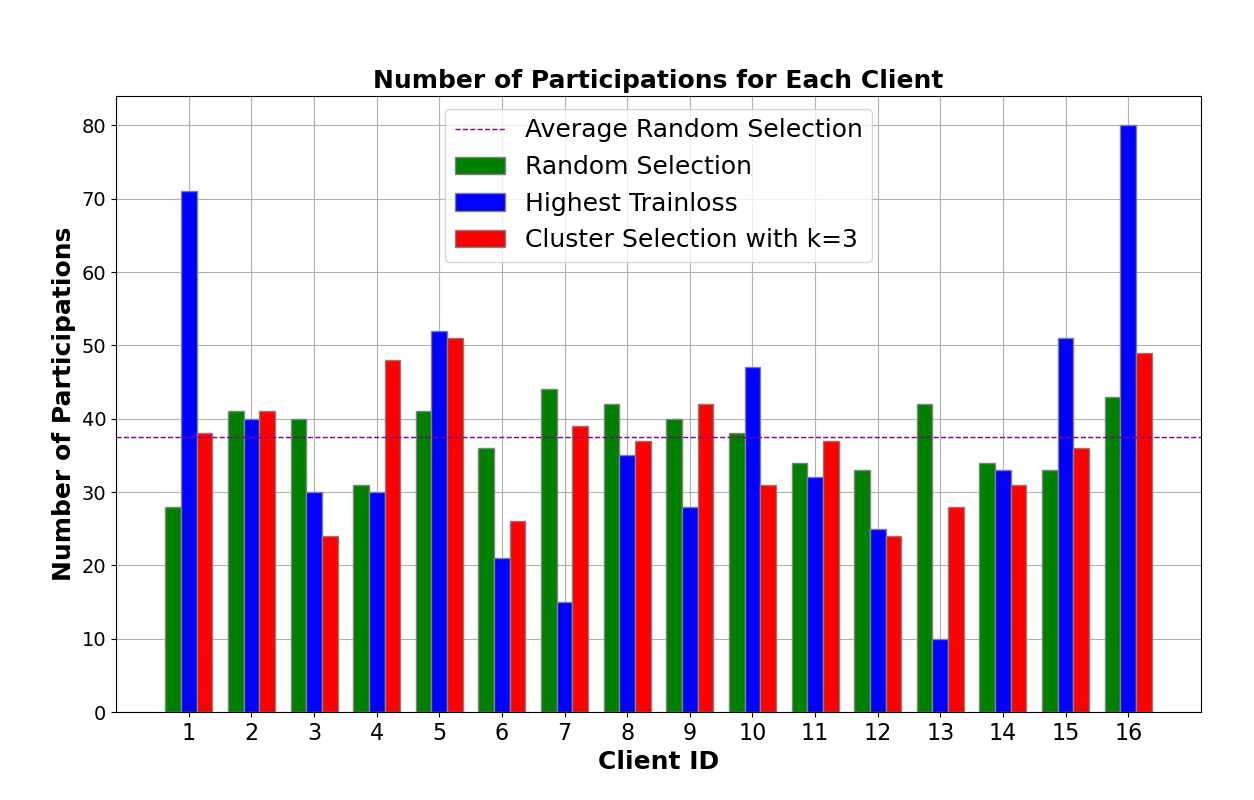}
    \caption{Comparison of client participation frequencies across different selection methods: The chart displays the number of participations for each client using three selection strategies: Random selection, highest training loss and cluster selection with k=3.}
    \label{fig:hist_selection}
\end{figure}

\begin{figure}
    \centering
    \includegraphics[width=1.0\linewidth]{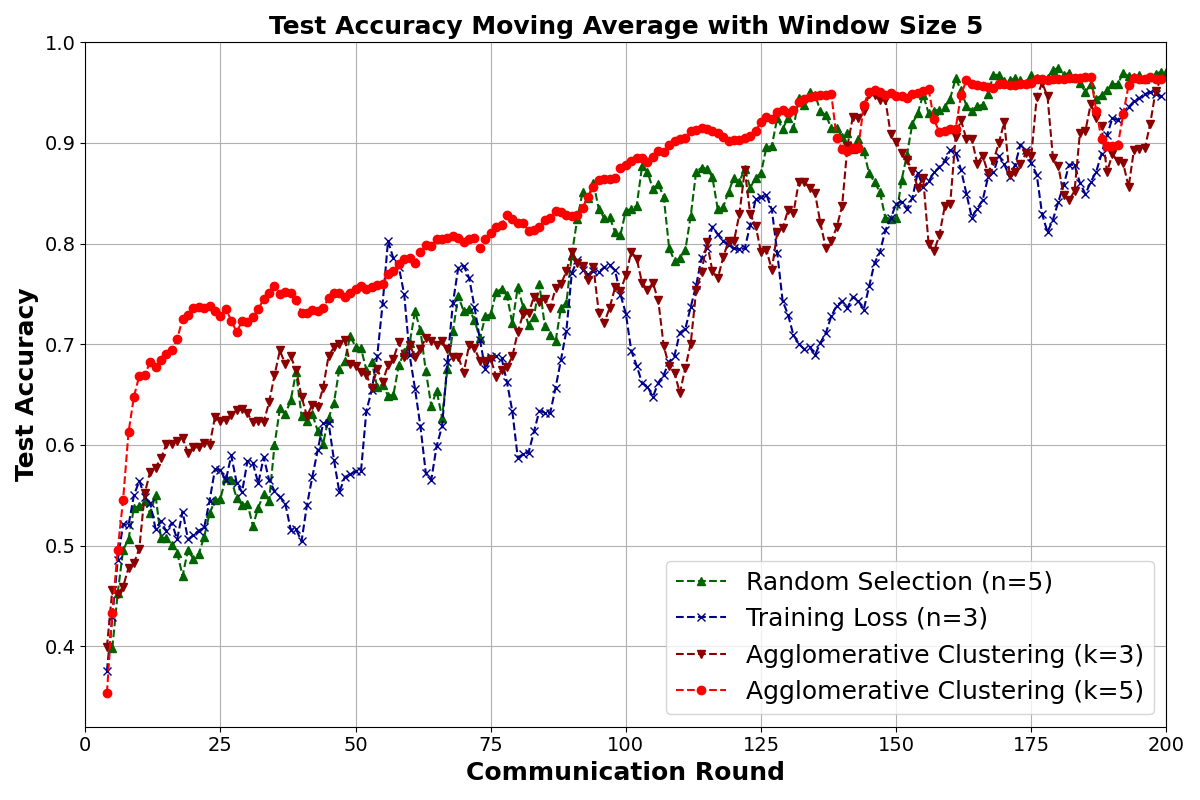}
    \caption{The plot shows the moving average of test accuracy across 200 communication rounds for three client selection methods: Random selection, training loss-based selection, and agglomerative clustering, each with their corresponding parameters.}
    \label{fig:test_acc_moving}
\end{figure}

To ensure that cluster-based client selection is effective beyond our dataset, we conducted further experiments using a subset of ImageNet. This subset includes three superclasses: animals, tools, and vehicles, each containing multiple subclasses (for instance, various classes for cats, dogs, and birds under the animals' superclass). Throughout all runs, the mapping of clients to datasets remained consistent, as did the randomization seed, such as when adding noise to the initially loaded ImageNet weights. The results, shown in Figure \ref{fig:test_acc_imgnet}, demonstrate that clustering strategies with any amount of clusters outperform the random and training loss-based strategies, particularly during the first 25 communication rounds. This analysis was conducted using the same hyperparameters as before, highlighting the effectiveness of clustering approaches in improving model accuracy, especially in the early stages of training, but also proofing that ImageNet is more complex and needs more fine-tuning to achieve a higher accuracy. Note, that clustering with k=3 outperforms random selection with n=5, meaning that in each communication round, two fewer clients are required to compute their updates, leading to a more efficient path to achieving high accuracy. It is important to highlight that the images within each superclass were re-clustered based on visual similarity using t-distributed stochastic neighbor embedding, leading to a non-IID distribution.
\begin{figure}
    \centering
    \includegraphics[width=1.0\linewidth]{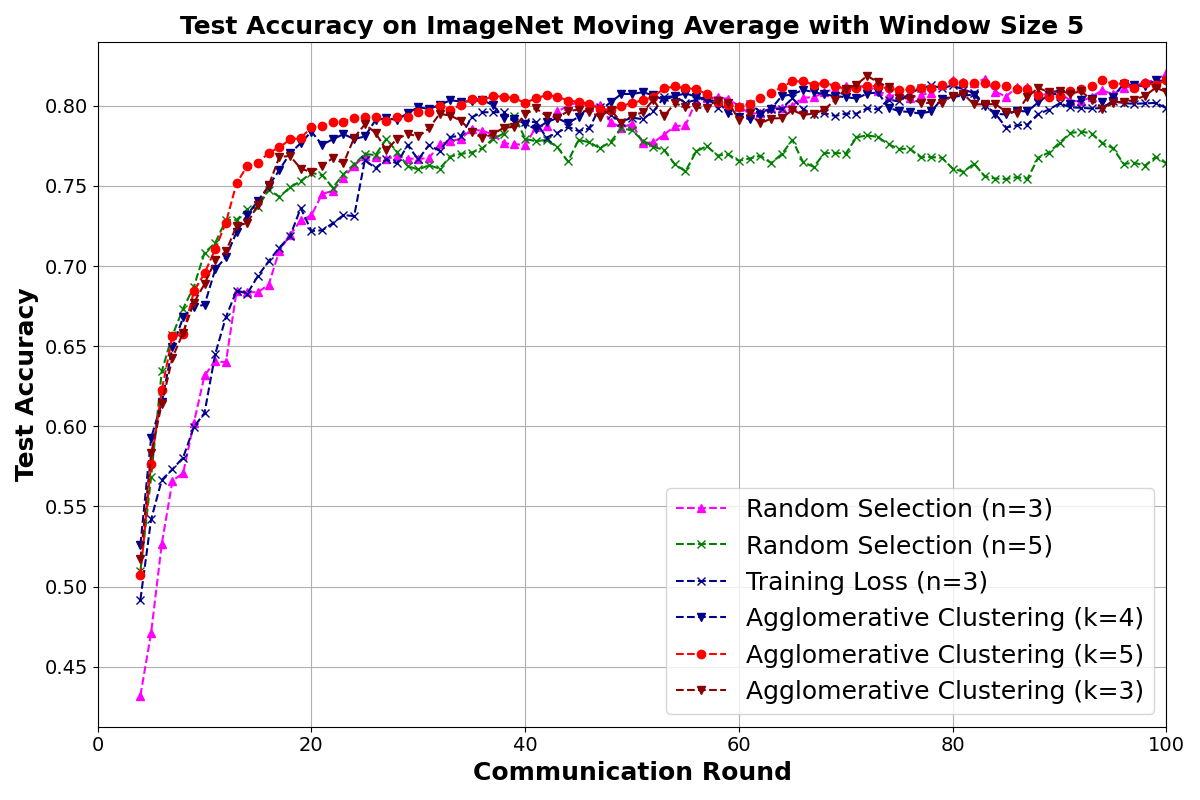}
    \caption{The plot shows the moving average of test accuracy on ImageNet across 100 communication rounds for three client selection methods: Random selection, training loss-based selection, and agglomerative clustering, each with their corresponding parameters. }
    \label{fig:test_acc_imgnet}
\end{figure}

\subsection{Computational Overhead}
Finally, we measured the client selection time on a local computer, as consistent hardware configurations cannot be guaranteed throughout all tests when using a high-performance cluster. When selecting 10 out of 16 clients for training—already a higher ratio compared to our other setups—the results indicate that agglomerative clustering is only slightly slower than random selection, with selection times of 0.0049 seconds and 0.0027 seconds, respectively. However, HDBSCAN requires significantly more time, taking 0.0444 seconds. Despite this, the communication time for transmitting weights in a network will far exceed these client selection times, especially in a cross-silo setup with a relatively low number of clients making the additional computation time for clustering methods less impactful in the overall process.

\section{Conclusion and Outlook}
In this study, we examined various client selection methods within the context of federated learning to address the challenges of integrating new clients with diverse and potentially non-IID data distributions. Our analysis demonstrated that clustering techniques, particularly agglomerative clustering, consistently outperformed random and loss-based client selection methods, offering both higher and more stable accuracy across communication rounds. This improvement is especially pronounced in the early stages of training, where efficient client selection plays a crucial role in accelerating the learning process while maintaining high model performance. The success of clustering-based methods underscores their potential in managing the complexities of data heterogeneity and enhancing the overall efficiency of FL systems.

Future research will focus on further refining clustering-based client selection strategies to better accommodate the dynamic nature of real-world FL environments, where clients may join or leave the system unpredictably. Additionally, the consideration of complex datasets like ImageNet, which has only been touched on in this study, is explored in greater depth in ongoing research and will be crucial to ensure their applicability across various domains. Another promising direction is the development of hybrid approaches that combine the strengths of different selection methods to optimize both accuracy and computational efficiency. As FL continues to evolve, addressing these challenges will be key to its successful adoption in production environments, particularly in industries where data privacy and diversity are paramount.

\section*{Acknowledgment}
This work was funded by the Carl Zeiss Stiftung, Germany under the Sustainable Embedded AI project (P2021-02-009).

\bibliographystyle{IEEEtran}
\bibliography{lit.bib}
\end{document}